\documentclass[letterpaper, 10 pt, conference]{IEEEtran}  
\IEEEoverridecommandlockouts
\usepackage{cite}
\usepackage{amsmath,amssymb,amsfonts}
\usepackage{algorithmic}
\usepackage{algorithm}
\usepackage{graphicx}
\usepackage{textcomp}
\usepackage{xcolor}
\usepackage{subcaption}
\usepackage{pdfpages}
\usepackage{bm}
\def\BibTeX{{\rm B\kern-.05em{\sc i\kern-.025em b}\kern-.08em
    T\kern-.1667em\lower.7ex\hbox{E}\kern-.125emX}}
\newcommand\numberthis{\addtocounter{equation}{1}\tag{\theequation}}
\usepackage[utf8]{inputenc}
\DeclareUnicodeCharacter{221E}{$_\infty$}

\begin{document}

\title{\vspace{6mm}
Counterfactual Explanation and Causal Inference In Service of Robustness in Robot Control
}



\author{\IEEEauthorblockN{Sim\'on C. Smith and Subramanian Ramamoorthy}
\IEEEauthorblockA{
\textit{Institute of Perception, Action and Behaviour} \\
\textit{School of Informatics}\\
\textit{The University of Edinburgh}\\
Edinburgh, UK \\
\{artificialsimon, s.ramamoorthy\}@ed.ac.uk}

\thanks{This work was supported by EPSRC funding for the ORCA Hub (EP/R026173/1, 2017-2021).}
}

\newcommand{\CLASSINPUTtoptextmargin}{19mm}
\newcommand{\CLASSINPUTbottomtextmargin}{18mm}
\newcommand{\CLASSINPUTinnersidemargin}{19mm}
\newcommand{\CLASSINPUToutersidemargin}{19mm}

\newcommand{\red}[1]{\textcolor{red}{#1}}

\maketitle

\begin{abstract}
We propose an architecture for training generative models of counterfactual conditionals of the form, \textit{`can we  modify event A to cause B instead of C?'}, motivated by applications in robot control. 
Using an `adversarial training' paradigm, an image-based deep neural network model is trained to produce small and realistic modifications to an original image in order to cause user-defined effects.
These modifications can be used in the design process of image-based robust control - to determine the ability of the controller to return to a working regime by modifications in the input space, rather than by adaptation. In contrast to conventional control design approaches, where robustness is quantified in terms of the ability to reject noise, we explore the space of counterfactuals that might cause a certain requirement to be violated, thus proposing an alternative model that might be more expressive in certain robotics applications.
So, we propose the generation of counterfactuals as an approach to explanation of black-box models and the envisioning of potential movement paths in autonomous robotic control. Firstly, we demonstrate this approach in a set of classification tasks, using the well known MNIST and CelebFaces Attributes datasets. Then, addressing multi-dimensional regression, we demonstrate our approach in a reaching task with a physical robot, and in a navigation task with a robot in a digital twin simulation.
\end{abstract}


\section{Introduction}

A robust autonomous system is one that can cope with perturbations and uncertainty in the state space~\cite{zhou1998essentials}. These systems are designed to maintain a specified level of performance despite disturbances in the input, or perturbations in parameters, typically assumed bounded by a well-defined set.
In realistic high-dimensional robotic systems, determining these limits tends to be highly non-trivial. In robotic systems that make use of machine learning methods such as deep neural networks (DNN), it is difficult to establish properties of behaviour outside the support of the distribution of the training data ~\cite{zhang2016understanding}. Moreover, even minimal carefully crafted disturbances (keeping the input close in statistical terms to the original data distribution), known as adversarial attacks, can already produce undesirable behaviour in DNNs~\cite{szegedy2013intriguing}. In applications involving persistently autonomous and unmanned vehicles, difficult configurations of the robot-environment may only appear after several hundreds or thousands of hours of normal functioning. So, training such autonomous systems to be robust calls for alternate approaches to characterising and utilising the bounds of normal operation. 

In the present work, we propose a generative model (called {\textit{cGen}}) that produces \emph{minimal} and \emph{realistic} counterfactual `disturbances' to a known input. The cGen counterfactual generator is used as a means to characterise robot control robustness. We do this via the level of modification to the environment that causes the controller to cease to be operational with respect to some requirement. In a sense, this model is used to compute the equivalent of stability boundaries for dynamical systems, but focussed here on issues arising from the perception-action loops.
By identifying such counterfactual modifications to a specific environment configuration where the controller fails to achieve the required goal, we are in a position to add to the training set additional configurations  \emph{hallucinated} by the generative model. We use the term hallucination as a way to describe the generation of data points that are close to the original distribution of the training data but are not explicitly present in them~\cite{salimans2016improved,heusel2017gans}. One use of such counterfactuals is as a way to explain classification decisions by DNNs. In this case, the counterfactuals make evident the required modification (disturbance) to an image to be classified in a user-defined category, differently from the original one. In contrast to work on adversarial attacks that focus attention on misclassification with imperceptibly small disturbances, our counterfactuals include a target class and close-to-the-original data distribution modifications.  

The training regime of cGen is based on the Generative Adversarial Network (GAN) approach~\cite{goodfellow2014generative}. For image classification, we train a generative model that takes input from the training data and modify it to be classified as a user-defined class. The generative model has two main objectives. Firstly, to modify the original input to fool the classifier (discriminator) to classify it as belonging to the target class. Secondly, the modification has to be as small a possible, i.e. the distance from the original and the modified image has to be minimal~\cite{wachter2017counterfactual}. The loss function of the generator is a linear combination of these two features. Following the usual GAN training pattern, we train the classifier to discriminate between unmodified target class images and the ones modified by the model. When addressing robot control regression problems, we include an extra module - the predictor (controller) of the system. The predictor takes the modified image (counterfactual) as input and produces real-valued output that can represent motor commands, plans, end-effector positions, etc. In this architecture for regression, the generator also includes the loss of the target controller in its loss function. The classifier now behaves closely to the role of the discriminator in a regular GAN, by discriminating between modified and non-modified images. Note that the predictor is fixed during all the steps of the cGen training. In this stage, we are focused on finding the optimal counterfactuals rather than improving the controller.

In causal inference, counterfactuals are related to the idea that ``if event A had not occurred, event B would not have occurred". Counterfactuals define a causal dependency between them and the outcome of applying it (or not) to the actual event. This is a more powerful definition than the correlation between cause and effect. Causal dependencies allow us to be certain that the counterfactual induces the results rather than mere co-occurrence. These characteristics make counterfactuals a better candidate to assess the robustness of a system compared to other utility functions like prediction accuracy or performance.

To evaluate this proposed approach, we study three cases. First, we use cGen in image classification tasks. In this case, the counterfactuals {\textit{explain}} the difference between the two classes. The induced disturbance makes explicit the features present in the image that the classifier uses to predict a specific category. In the second case, we study the causal {\textit{modification}} needed to solve a robot control problem. Two experiments are carried out, one on a physical PR2 robot with an end-effector planning problem, and the other a Model Predictive Control (MPC) problem involving a simulated Husky robot. In the third case, we study our proposed counterfactual robustness measure between MPC controllers with different degrees of generalisation.

\section{Related Work}

Methods for robust controller design, such as $H^\infty$ loop-shaping~\cite{glover1992loop},  have been successfully used to design robust controllers for autonomous robotic control~\cite{la2002design}, manipulators~\cite{yeon2008practical} and navigation~\cite{rigatos2015new}, where external perturbations are suppressed by weighting the system transfer function appropriately. Similarly, sliding mode control~\cite{zinober1989deterministic} has been used with robotic manipulators and navigation~\cite{yu2005continuous,wang2009neural,guldner1995sliding}, in order to have the system \emph{slide} along a suitable sub-manifold defined by multiple discrete control modes. More foundationally, the notion of Lyapunov stability based control~\cite{zinober1994variable}, has been used also in multi-agent coordination~\cite{ogren2001control}, path following~\cite{park2007performance} and hybrid models of  bipedal walking~\cite{ames2014rapidly}, \cite{ramamoorthy2006qualitative}, where the system is guaranteed to stay near an equilibrium point if the solution is near the equilibrium.

In control theory, one uses notions such as gain/phase margin and the combination of both of them as disk margin to measure the robustness of such controllers.
These measurements are typically defined in a frequency domain associated with the single-input single-output (SISO) or  multiple-input multiple-output (MIMO) systems' feedback loop~\cite{blight1994practical}. These margins indicate the limits of disturbances before the system loses maintenance of performance criteria or even stability. Thus, in general, controllers with larger margins are  more robust than controllers with smaller margins. Given detailed models and when we have access to models such as the transfer function, we can derive these margins numerically or graphically.

However, such approaches are not well suited to the design of adaptive systems such as those using DNN-based control~\cite{chu1999tuning}. Here, we propose a model-agnostic measurement approach for robustness based on the distance from a perturbed state to the closest state within the working limits of a robotic visually grounded controller. Even in data-driven systems, the sought state may not exist in the data. Thus, we use a generative model for data hallucination.

Data hallucination or `imagination' relates to the generation of data points that are plausible in the domain, but are not part of the original training data set. For example, in image generation GAN methods, a generative model is able to produce images that are indistinguishable from the original set by a discriminator. Conditional-GAN and style transfer~\cite{sricharan2017semi,gatys2016image} are examples of image generation based on feature transformation. Usually, GAN-based methods measure the quality of the generator by computing the entropy of the conditional probability of the generated image belonging to the training set and the entropy of the marginal probability of the generated images~\cite{salimans2016improved,heusel2017gans}. Along the same lines, modification of the latent space in Variational Autoencoders (VAE)~\cite{kingma2013auto} has been studied to interpolate features in images in combinations that do not appear in the training data~\cite{mathieu2016disentangling,berthelot2018understanding}. For control tasks, the authors in~\cite{wu2019imagine} use data hallucination by modifying the latent space in a VAE to find tools that can be used to reach a goal in 3D space. Within Reinforcement Learning (RL), data hallucination has been used as a way to find plausible future states of the environment to train agents without having to take the action in the real environment~\cite{NIPS2017_7152,kalweit2017uncertainty} and for imagining predictable and sensitive future states~\cite{10.1007/978-3-642-28258-4_9}. Differently from model-based RL, where a forward model is used to predict the next state based on the actual state and action, this class of RL methods use internal models to  generate rollouts where an imagined reward is used to improve the policy. In regret based strategies~\cite{lee2018dynamic}, an agent compares the actual action to all the other actions that did not take, answering ``what could be the value if the agent had tried another action?''.

Causal inference enables determination of cause-effect connections~\cite{pearl2009causal} in order to be able to make explainable predictions about the world. Causal models provide a framework for modelling causation in addition to statistical relationships~\cite{pearl2009causality,nair2019causal}. A useful representation in causal models is that of a directed acyclic graph representing the cause of an effect as its parent nodes~\cite{NIPS2008_3548,shajarisales2015telling}.
The richer structure of causal learning compared to statistical learning allows for a better understanding of the underlying cause-effect properties of a system~\cite{peters2017elements}.
This deeper understanding is exploited to explain the behaviour of black-box models~\cite{alvarez2017causal,singla2019explanation} and for interpretability~\cite{kim2017interpretable}.
As a specific case of causal inference, counterfactuals have been used to study fairness, accountability and transparency in machine learning~\cite{wachter2017counterfactual,van2019interpretable}. In these cases, the authors use tabular data and low-dimensionality images, where a random search is performed directly over the state variables or over a set of features. In high-dimensional data streams such as with RGB images, counterfactual analysis has been used to explain classifiers by taking salient blobs of a target class image and placing it in the original class image~\cite{goyal2019counterfactual}; In~\cite{chang2018explaining}, an in-filling blob is generated following the original data distribution to change the classification of a DNN. Most of these approaches represent counterfactuals as part of an optimisation problem for a particular instance of the state. In our approach, we train a generative model to produce the counterfactuals for any arbitrary data point.


\section{Counterfactual Generator}

We use counterfactual inference in three different use cases. First, in the case of a visual robot control, we use it to find the smallest required modification to the input state that allows the robot to solve a control task. Consider an image-based controller under adversarial conditions that prevent the robot from reaching its objective. In this scenario, the desired counterfactual will be the minimal and realistic modification to the image (scene) that allows the robot to indeed achieve the task. For a realistic modification, we mean a change in the state that could plausibly be performed in the real environment. For example, adversarial attacks can be a significant distribution change in the input, but imperceptible  to the human eye~\cite{szegedy2013intriguing}. On the contrary, the counterfactuals that we consider are changes that have a meaning in the domain, e.g. the configuration of obstacles in a navigation task.  Our proposed method is able to generate a solution based on the actual input state, and to provide an evaluation of the efficiency of the solution. We define the relative robustness of a model with respect to another by comparing the distance from the original state to the counterfactually modified state. A lesser distance implies that the counterfactual induces a smaller modification of the original input. Thus, a smaller counterfactual distance implies that the controller is more efficient to recover from adversarial attacks.
The generation of counterfactuals is analogous to adaptive methods in configuration space~\cite{bongard2006resilient}. In these approaches, the robot imagines a solution, via an internal model, for an adversarial attack and then applies the solution to its own configuration. In our case, the counterfactual is in the state space, where an external stakeholder has to apply the modification to the environment.

Now, we present the architecture of our counterfactual generative model for classification model explanation. In this approach, we want to find the minimal modification to the input that will make the classifier change the prediction from the original class to a target class. Given an input $\bm x \in \mathbb{R}^n$ with $n$ dimensions, we want to find an $\bm{x'} \in \mathbb{R}^n$ that is the closest to the original input that has the highest probability of belonging to the target class $t_c \in \mathbb{N}$:
\begin{equation}
  \min_{\bm{x'}} d_g(\bm x, \bm{x'}) + d_c(C(\bm{x'}), t_c),
  \label{eq:original_counterfactual}
\end{equation}
 where $d_g \colon \mathbb{R}^n \times \mathbb{R}^n \to \mathbb{R}$ and $d_c \colon \mathbb{R} \times \mathbb{R} \to \mathbb{R}$ are distance functions on the input space and in the class space respectively, and $C \colon \mathbb{R}^n \to \mathbb{R}$ predicts the probability  of the input belonging to $t_c$.

 
 We propose a generator model $G \colon \mathbb{R}^n \to \mathbb{R}^n$ that takes any arbitrary input and generates the modified version of the input:
\begin{equation}
  \bm{x'} = G(\bm x, \bm{\theta}) \\
\label{eq:generator}
\end{equation}
where $\bm{\theta} \in \mathbb{R}^m$ represents the $m$ parameters of the generator that are trained to minimising the cGen loss function $L_{\text{cgen}}$:
\begin{align*}
  L_g(\bm x, \bm{x'}) &= d_g(\bm x, \bm{x'}) \\
  L_c(\bm x, t_c) &= d_c(C(\bm x), t_c) \\
    L_{\text{cgen}}(\bm x, \bm{x'}) &= L_g(\bm x, \bm{x'}) + L_c(\bm x, t_c) \numberthis
   \label{eq:loss_discriminator}
\end{align*}

\begin{figure}
  \includegraphics[width=\linewidth]{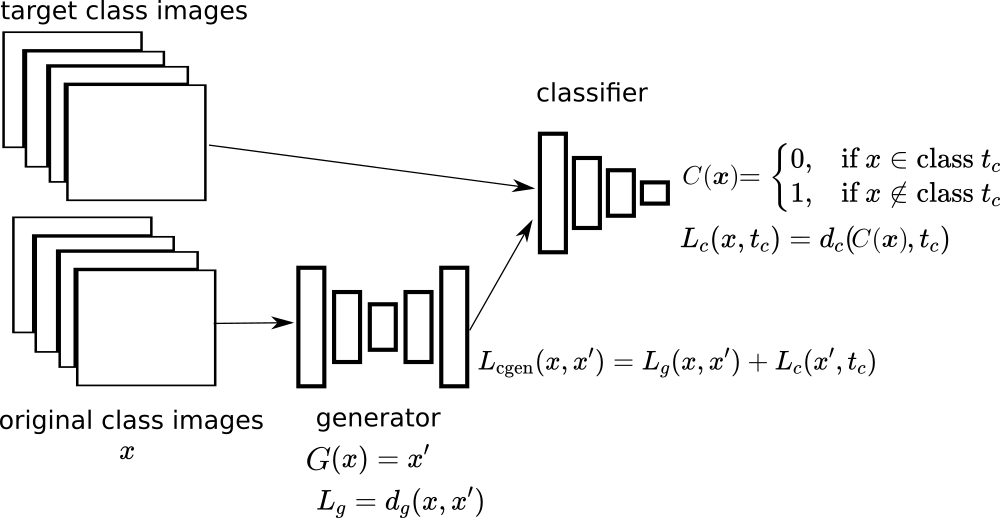}
  \caption{The cGen architecture for classification. The generator and the classifier are implemented as deep convolutional neural networks.}
  \label{fig:cgen_classification}
\end{figure}

Following the family of generators GAN, we replace the GAN generator for a counterfactual generator with an auto-encoder architecture. The counterfactual generator optimises Eq.~\ref{eq:loss_discriminator} rather than the usual reconstruction loss.  Fig.~\ref{fig:cgen_classification} shows the architecture of the classification cGen. Compared to a regular GAN, the cGen takes as input the original value $\bm x$ rather than a randomised input sampled from a probabilistic latent space. The binary classifier is trained to predict if $\bm x$ belongs to the target class $t_c$. Since the generator has the same architecture as an auto-encoder, we can pre-train it with the original class data subset and a reconstruction loss.

We train the cGen with a GAN mechanism, optimising the generator and the classifier in consecutive epochs. The generator is trained to find the parameters that minimise the weighted multi-variable loss from Eq.~\ref{eq:loss_discriminator} as:
\begin{equation}
  L_{\text{cgen}}(\bm x, \bm{x'}) = (1 - \alpha) L_g(\bm x, \bm{x'}) + \alpha L_c(\bm{x'}, t_c),
\label{eq:loss_discriminator_weighted}
\end{equation}
with $\alpha \in [0, 1]$. 
The first term on the RHS is  the generator loss. The second term on the RHS represents the classifier loss.
The idea is that the generator preserves as much as possible the original input (as an auto-encoder does), but also modifying the input, so it is classified as the target class.
A value of $\alpha = 1$ generates counterfactuals closer to the target class without any regard for the original input. A value of $\alpha = 0$ will generate counterfactuals identical to the original input without considering the target class.
 The weights of the classifier are frozen during the training of the generator. On alternating epochs, the classifier is trained to discriminate images belonging to the target class data set or from the generator.

\subsection{Classification results}
We test the classification version of the cGen with the MNIST handwritten digits
and celebrity faces dataset~\cite{liu2018large}.
\subsubsection{MNIST}

We implement the classifier, encoder and decoder part of the generator with four convolutional/deconvolutional layers, and a fully-connected output layer with sigmoid activation. The distance functions $d_g$ and $d_c$ (Eq.~\ref{eq:loss_discriminator}) were implemented as mean squared error.
We pre-trained the generator with the original class subset and the classifier with the training data set from MNIST as a binary classifier for class number 8.

\begin{figure}
  \includegraphics[width=\linewidth]{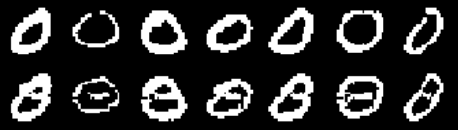}
  \caption{The top row shows MNIST number in the original class 0. The bottom row shows the counterfactuals for target class 8. Note how the shape of the 0 is maintained in the counterfactual. Balance weight $\alpha = 0.8$.}
  \label{fig:results_mnist}
\end{figure}

Fig.~\ref{fig:results_mnist} shows a set of results after the training phase for the original class number 0. The top row shows original class images, and in the same column on the bottom row, the generated output for each image. The examples shown are taken from the validation set, not seen during training. Note that the counterfactuals were obtained from the same generator, on a forward pass for each original image.

\subsubsection{CelebA}

For the celebrities faces dataset, we use a similar implementation adjusting input and kernel sizes from the MNIST implementation. We test the counterfactual generator with two experiments. One for the original class ``no sunglasses'' and target class ``sunglasses''. The other experiment is for classes ``no smile'' and ``smile''. Following the same steps as with the previous dataset, after pre-training the generator, we run the cGen optimisation.

\begin{figure}
  \centering
  \begin{subfigure}{0.45\textwidth}
    \centering
    \includegraphics[width=0.9\linewidth]{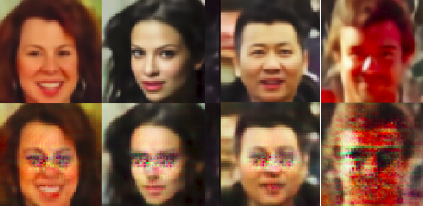} 
    \caption{Sunglasses class}
  \end{subfigure}
  \begin{subfigure}{0.45\textwidth}
    \centering
    \includegraphics[width=0.9\linewidth]{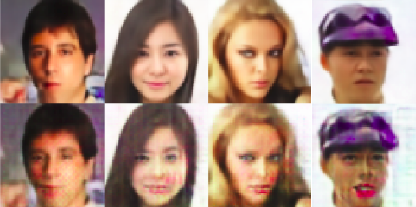} 
    \caption{Smile class}
  \end{subfigure}
  \caption{The top row shows the  images of celebrities in the original class. The bottom row shows the counterfactuals for the target class. $\alpha=0.8$.}
  \label{fig:celeba}
\end{figure}

Fig.~\ref{fig:celeba} shows the results for the celebrity faces experiment. For the sunglasses category, the counterfactuals (bottom row) show noise in the position where sunglasses are usually located. The last example in the figure shows two black circles in the eyes positions, but also adds noise to the rest of the face. For the smile category, more subtle modifications are made to the lips and the area surrounding it.

\section{cGen for Regression}
With a view to applying this methodology to robot control tasks, we now consider the use of the cGen  for real-valued predictions.

\begin{figure}
  \includegraphics[width=\linewidth]{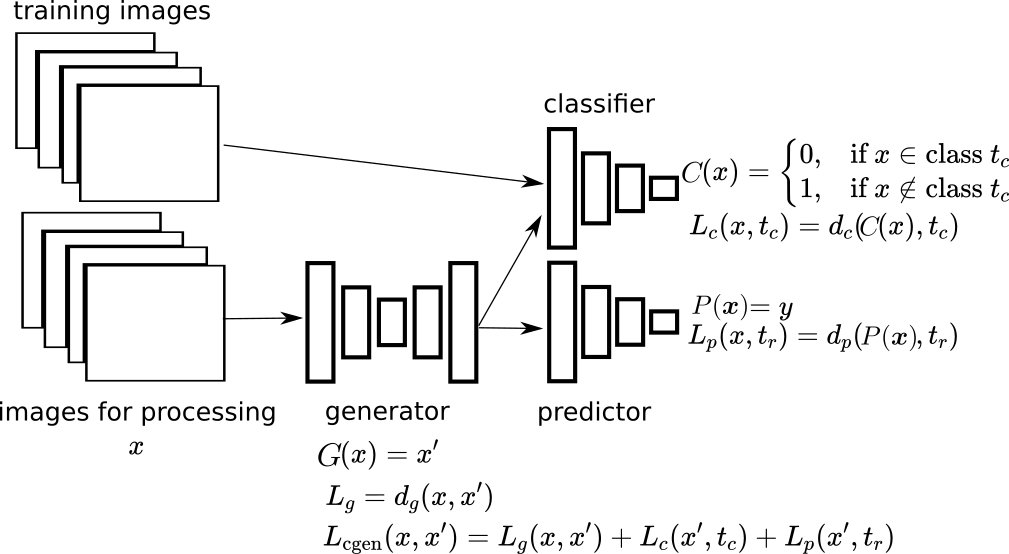}
  \caption{Regression cGen architecture. The target class $t_c$ of the classifier corresponds to $\bm x$ being part of the training dataset. The predictor includes a target regression $t_r$.}
  \label{fig:cgen_regression}
\end{figure}

We add a new component to the cGen architecture. The new component is a predictor function $P \colon \mathbb{R}^n \to \mathbb{R}^m$, with $m$ as the prediction dimension. Fig.~\ref{fig:cgen_regression} shows this architecture. In this setup, the generator includes a new component in the $L_{\text{cgen}}$ loss function, the predictor loss $L_p$. The role of the predictor is to evaluate the counterfactual with respect to a goal prediction. Now, the classifier acts like a regular GAN discriminator. The classifier is used to keep the output of the generator as close as possible to training data set distribution. Minimising the classifier loss accounts for a coherent counterfactual with respect to the training data.

The predictor is the model that calculates the regression. For example, the predictor can be the policy of an agent, a weather forecast, or any type of real-valued function. The new component of the cGen loss captures the prediction error:
\begin{equation}
  L_p(\bm{x'}, t_r) = d_p(P(\bm{x'}), t_r),
  \label{eq:loss_prediction}
\end{equation}
where $P(\cdot)$ is the predictor model, $t_r \in \mathbb{R}^m$ a target value and $d_p \colon \mathbb{R}^m \times \mathbb{R}^m \to \mathbb{R}$ a distance function between regression targets. The new cGen loss function for regression is:
\begin{equation}
  L_{\text{cgen}}(\bm x, \bm{x'}) = \alpha L_g(\bm x, \bm{x'}) + \beta L_c(\bm{x'}, t_c) + \gamma L_p(\bm{x'}, t_r).
  \label{eq:loss_cgen_regression}
\end{equation}
Note that $t_c$ is not a target class as in the classification case. Now, $t_c$ targets whether $\bm x$ is part of the training data.

\subsection{Counterfactuals for Robot Control}
\begin{figure}
  \centering
  \begin{subfigure}{0.48\linewidth}
    \centering
    \includegraphics[width=0.99\linewidth]{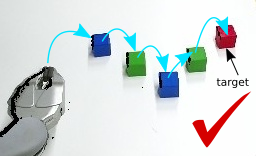} 
    \caption{Successful control}
    \label{fig:pr2_control_a}
  \end{subfigure}
  \hfill{}
  \begin{subfigure}{0.48\linewidth}
    \centering
    \includegraphics[width=0.99\linewidth]{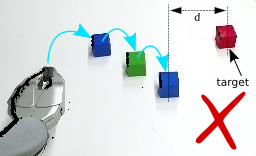} 
    \caption{Failed control}
    \label{fig:pr2_control_b}
  \end{subfigure}
  \caption{Image (a) shows a successful scenario where the end effector can reach the target object. The cyan arrows represent the path that the end effector will follow to reach the target (red object). Image (b) shows a scenario where the controller is unable to find a path between the green and red object as the distance $d$  between them is larger than the $\delta$ threshold.}
\end{figure}

To test the regression cGen architecture, we implement autonomous control on a PR2 robot. Fig.~\ref{fig:pr2_control_a} shows the movement of the end effector of the left arm of the robot. From an initial position on the left side of the scene, the end effector transits over the objects to arrive at the red object. The movement from one object to the next is conditional on the distance to the next object. A movement can only be performed if a distance $d$ to the next object is smaller than a set threshold $\delta$. If the distance $d$ to the next object is larger than the threshold $\delta$, the end effector is unable to arrive at the target, and has to remain in the same position failing the objective. This restriction can be seen as a safety measure or domain-specific constraint. Fig.~\ref{fig:pr2_control_b} shows a failed execution as the distance from the last visited object to the red cube is larger than $\delta$. We would like to find a counterfactual, i.e. a minimal and realistic modification to the image, wherein the robot does indeed arrive at the target. Valid solutions can include the adjustment of the position of an object, e.g. move the target closer, or adding new objects to fill in the gap between objects. 

We implement the visual processing based on standard (and fairly classical) computer vision techniques, including thresholding and edge detection. However, a controlled based on this input is not differentiable. Thus, we are not able to use it directly as the predictor in the cGen architecture. To overcome this problem, we trained a neural network with data pairs generated by the original controller. In this network, the input is the image used by the controller, and the output is the distance of the last achievable position of the end effector. If the end effector reaches the target, the output is 0. If the end effector does not reach the target, the output is the distance between the last position of the end effector and the next object.  We collected image samples from the on-board camera of the robot. Using data augmentation techniques, we trained a convolutional DNN as a differentiable approximator of the original controller.

Similar to the classification cGen, we can pre-train the generator, but not the classifier (as now it depends on generated images). We use adversarial training for cGen, minimising Eq.~\ref{eq:loss_cgen_regression}, with fixed weights for the predictor and the classifier. After one epoch training the generator, we train the classifier with real images and images from the generator. After a short iterative search, the meta-parameters where set to $\alpha=0.8$, $\beta=0.1$ and $\gamma=0.1$. The final results from the generator are passed trough an auto-encoder trained on the original images. This step reduces the noise in the counterfactual making it easier to compare to the original image.

\begin{figure}
  \centering
  \begin{subfigure}{0.31\linewidth}
    \centering
    \includegraphics[width=0.99\linewidth]{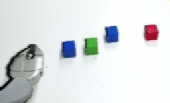} 
    \caption{Original}
    \label{fig:pr2_result_a}
  \end{subfigure}
  \hfill{}
  \begin{subfigure}{0.31\linewidth}
    \centering
    \includegraphics[width=0.99\linewidth]{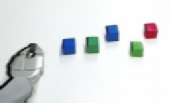} 
    \caption{Counterfactual}
    \label{fig:pr2_result_b}
  \end{subfigure}
  \hfill{}
  \begin{subfigure}{0.31\linewidth}
    \centering
    \includegraphics[width=0.99\linewidth]{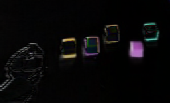} 
    \caption{Difference}
    \label{fig:pr2_result_c}
  \end{subfigure}
  \caption{(a) is the original input of a failed controller. The end effector is only able to move to the third object (from left to right), missing the target. (b) The counterfactual shows the addition of a new object to the scene. With that modification, the controller can reach the target. (c) is the difference between the original image and the counterfactual. The pink blob represents the new object in the scene. Some reconstruction noise is present around the objects in the original image.}
  \label{fig:pr2_result}
\end{figure}

Fig.~\ref{fig:pr2_result_a} shows the original image of a failed controller where the robot is unable to reach the target object (red cube). Fig.~\ref{fig:pr2_result_b} shows how the generator adds an object (green cube) in the gap between the target (red cube) and the previous object (blue cube). Fig.~\ref{fig:pr2_result_c} highlights the difference between the two images. Fig.~\ref{fig:pr2_result_b}~and~\ref{fig:pr2_result_c} serve as an explanation of what modification to the original scene are needed to have a successful controller. Now, we can configure the scene following the counterfactual (adding the green cube) and run the controller. With this modification, the end effector is able to reach the target (not shown on the images).

\begin{figure}
  \includegraphics[width=\linewidth]{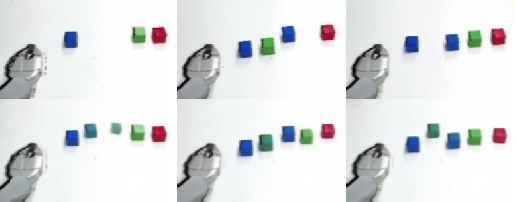}
  \caption{Other results for regression cGen. The top row shows original images, counterfactuals shown in the bottom row. The generator can add more than one object in the scene, so the controller arrives at the goal. Also, objects are added in other positions to close the gap.}
  \label{fig:pr2_more_results}
\end{figure}

Fig.~\ref{fig:pr2_more_results} shows more counterfactuals. The top rows are the original scenarios where the controller fails. The bottom rows are the counterfactuals. Note that in the first scenario, the counterfactual adds two objects to fill the gap. The counterfactual were executed by the controller successfully.

\subsection{Model Predictive Control}
\label{sec:mpc}

\begin{figure}
  \centering
  \begin{subfigure}{0.22\columnwidth}
    \includegraphics[width=\columnwidth]{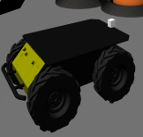} 
    \caption{ Husky }
  \end{subfigure}
  \hfill
  \begin{subfigure}{0.22\columnwidth}
    \includegraphics[width=\columnwidth]{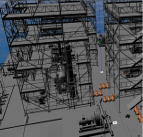} 
    \caption{Oil rig}
  \end{subfigure}
  \hfill
  \begin{subfigure}{0.24\columnwidth}
    \includegraphics[width=\columnwidth]{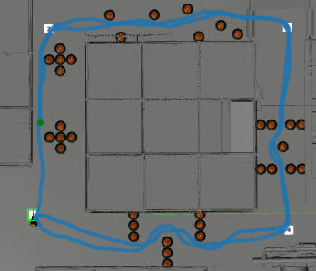} 
    \caption{User demo}
  \end{subfigure}
  \hfill
  \begin{subfigure}{0.20\columnwidth}
    \includegraphics[width=\columnwidth]{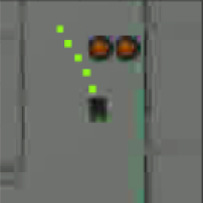} 
    \caption{MPC}
  \end{subfigure}
  \caption{Learning from demonstration: A user tele-operates a Husky robot (a) on an oil platform digital twin (b) for a surveillance and inspection task (c). The user navigates the robot around the four corners of the platform (blue lines). The system learns from a local top-view camera centred over the robot (d) the future position of the robot (yellow dots).}
  \label{fig:lfd}
\end{figure}

Now, we test our approach in a high-dimensional regression robotic control problem. In a Learning from Demonstration (LfD)  setup, we train a controller to predict the future position of a tele-operated robot in a surveillance task (Fig.~\ref{fig:lfd}). The controller learns to predict the future positions in a set of demonstrations provided by the operator. The task for the operator is stated as: starting from the bottom-left corner (Fig.~\ref{fig:lfd}c), visit all the corners of the platform in a clockwise direction and return to the starting point. The future position that the controller has to learn to predict is a 10-dimensional vector with the $x$- and $y$-axis position of five pairs in the next $5$ seconds in intervals of $1$ second (Fig.~\ref{fig:lfd}d, yellow dots). After training, the learned controller can be used in a model predictive control paradigm to autonomously control the robot. 

For this experiment, we modify the cGen architecture for regression presented in the previous section. We modify the generator model from an autoencoder to its variational counterpart~\cite{kingma2013auto}. With this modification, we can directly optimise the generator over its latent space $Z$ rather than modify all the parameters of the neural network. First, we pre-train the generator to minimise the reconstruction error. Then, we can directly optimise the cGen loss (Eq.~\ref{eq:loss_cgen_regression}) with stochastic gradient descend over the latent variables $z_i \in Z$, with $i = \{1, 2, \dots, l\}$ and $l$ the number of latent variables.
This modification of the architecture has two main advantages. First, the optimisation process of the generator is linear with respect to the size of the latent space ($\mathcal{O} (l)$), thus speeding up the training time. Second, the counterfactual generated images are an interpolation of the latent space, rendering images closer to the original training set. The downside of this architecture change  is that the decoder and encoder modules of the generator need to be pre-trained and the counterfactual search has to be performed over each instance of an  input image $\bm x$, and one regression goal $t_r$.

\begin{figure}
  \centering
  \includegraphics[width=0.8\columnwidth]{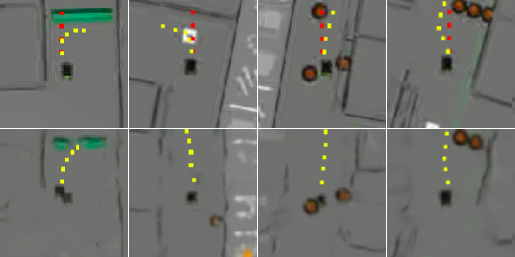}
  \caption{Counterfactuals generated by the cGen architecture. The top row shows the original input with the model predictive control in yellow dots and the goal in red dots. The bottom row shows the counterfactuals and their predictions. }
  \label{fig:orca_vae}
\end{figure}

Fig.~\ref{fig:orca_vae} shows counterfactuals for a set of scenarios taken from the demonstrations with a goal that is different from the prediction of the controller. The top row shows the original scenarios with the model predictive control as the yellow dots. The straight line of red dots represents the regression goal for the counterfactuals $t_r$. The bottom row shows the counterfactuals for the original input. In the first figure, the generator changes the green barrier for two barriers with a separation in the middle. Note the counterfactual predictive control (bottom row yellow dots) is closer to the goal (top row red dots), but is not an exact solution for the required goal. This difference is due to the fact that there is no scenario in the demonstration that includes a green barrier only on the right side of the hallway. Such a solution is discarded by a higher loss in the discriminator. For the rest of the scenarios, the counterfactuals include addition and removal of obstacles and rotation of the scene. 

\section{CGen for Robustness Analysis}

As discussed earlier, robust control design considers how well a controller maintains performance criteria when facing external perturbations. One controller is considered to be more robust than another if it can control the plant under larger perturbations than the other controller.
In this section, we use counterfactual analysis to obtain an indicator of the efficiency of a controller to return to a working regime after it is affected by perturbations. The idea is that when a system faces a perturbation or an adversarial attack, then instead of adapting, e.g. via changes in the configuration space, the counterfactual analysis is able to find a solution in the state space. For example in the oil platform scenario, where the MPC is based on images taken from a top-view camera, a failure to arrive at a destination goal can be due to obstacles in the path of the robot. This scenario can be solved by  manipulating  the objects in the scene (not necessarily by the robot, instead by requesting human help). In this case, for different controllers, we can measure the quality of the counterfactual solution by the number of modifications to the scene induced by the cGen and by the distance between the desired goal and the one obtained with the counterfactual image. 

To measure the difference between controllers, we trained three controllers under different demonstrations. For the three controllers, the surveillance task remains the same as in Sec.~\ref{sec:mpc}, but with different obstacle complexity. The first controller (a) is trained in scenarios with two types of obstacles: orange cones and green barriers. The second controller (b) is trained from demonstrations with scenarios with cones but no barriers. The third controller (c) is trained in scenarios without obstacles. We define a set of goals for the MPC to predict movements of the robot in a range of $[-45, 45]$ degrees from the vertical line.

\begin{figure}
  \centering
  \includegraphics[width=0.8\columnwidth]{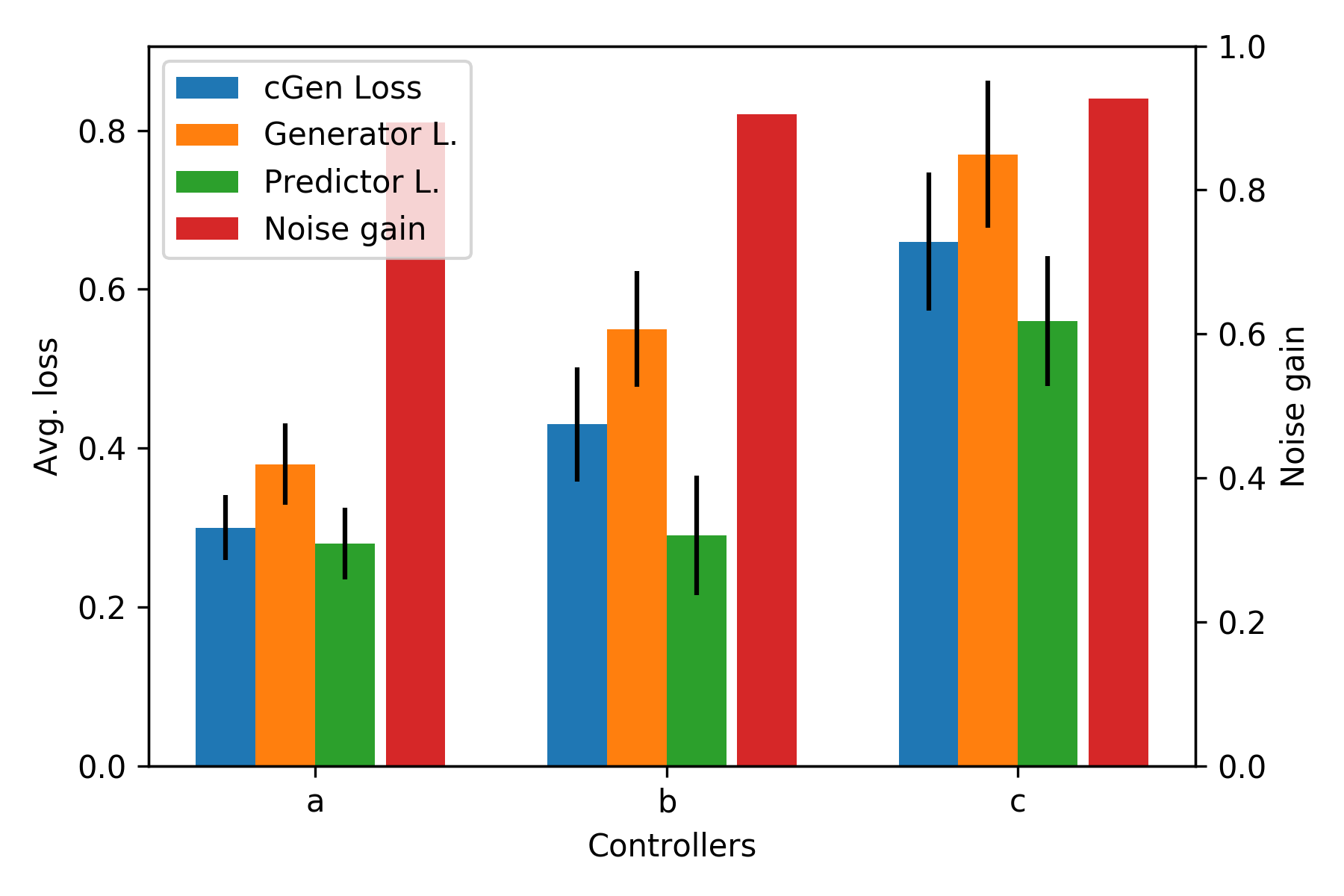}
  \caption{Three controllers trained under different levels of complexity of demonstration. All controllers have a similar response to perturbations ($\eta$ as the noise gain and $\varepsilon=0.1$). Controller \emph a was trained with full demonstrations. Controller \emph b worst performance compared to controller \emph a is explained by the unseen obstacles during training. Controller \emph c has the worst performance among the controllers. The lack of obstacles during training does not allow the controller to produce more complex behaviour.
  }
  \label{fig:robustness_graph}
\end{figure}

\begin{figure}
  \centering
  \rotatebox[origin=c]{0}{(a)}\quad
  \begin{subfigure}{.91\columnwidth}
    \includegraphics[width=\columnwidth]{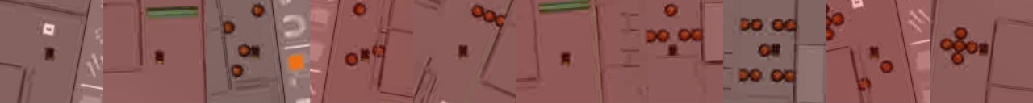} 
  \end{subfigure}
  \\
  \rotatebox[origin=c]{0}{(b)}\quad
  \begin{subfigure}{.91\columnwidth}
    \includegraphics[width=\columnwidth]{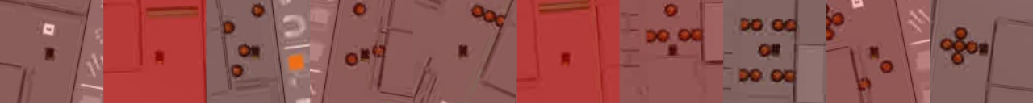} 
  \end{subfigure}
  \\
  \rotatebox[origin=c]{0}{(c)}\quad
  \begin{subfigure}{.91\columnwidth}
    \includegraphics[width=\columnwidth]{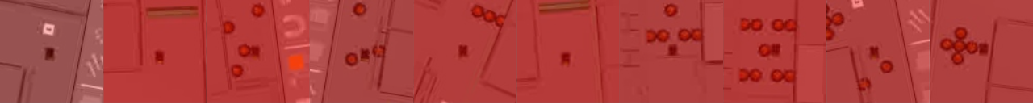} 
  \end{subfigure}
  \caption{Performance of the controller in different scenarios. Strength of the red colour indicates a higher counterfactual loss. Controller \emph b has a larger loss on scenarios with a barrier. Controller \emph c performs comparatively bad in all the scenarios.}
  \label{fig:robustness}
\end{figure}

As a baseline, we measure the robustness of the controllers to random noise in the input data. We inject noise $\mu$ with an increasing gain $\eta$ until the difference between the non-modified prediction and the modified one is larger than a set value $\varepsilon$: $|| p(\bm x) - p(\bm x + \eta \mu) ||^2 < \varepsilon$. 
If the difference is larger than $\varepsilon$, the controller has arrived to its limit to reject perturbations. 
Fig.~\ref{fig:robustness_graph} shows that the sensitivity to noise is similar for the three controllers, i.e. they all share similar robustness to noise perturbation. To measure the average size of counterfactual modifications required by each controller to return to a working regime, we use cGen to generate counterfactuals over a set of 10 scenarios with different goals. Fig.~\ref{fig:robustness} shows the average losses for the three controllers. The counterfactual cGen loss (Eq.~\ref{eq:loss_cgen_regression}) shows that controller \emph{a} has the lowest loss among the three controllers. This is confirmed by the generator and predictor loss. These two low values indicate that the counterfactuals are close to the original image and also produce a prediction close to the goals. Controller \emph{b} shows a worst generator loss but a similar prediction loss compared to controller \emph{a}. This behaviour can be explained by the missing training data including the green barrier, but still maintaining a good level of sensitivity to the input. The counterfactuals for the controller \emph{b}, when faced in a scenario with previously unseen obstacles (barriers), result in larger modification of the scene to allow the model to make a closer prediction to the goal. In this case, the counterfactual not only have to modify (or remove) the barrier; it also has to add cones to guide the controller to the goal. Controller \emph{c} shows the worst counterfactual performance among the three controllers. The generator and prediction loss are both high. We explain this behaviour as that the controller does not respond to obstacles on the input image --as it did not experience them during training--. Also, the counterfactuals will tend to remove all the object from the scene rather than applying smaller modifications.

Fig.~\ref{fig:robustness} shows the counterfactual loss for each scenario for controller \emph a, \emph b and \emph c. In the image, a stronger red colour means a higher value average loss for all the  goals. Comparing controller \emph a and \emph b, we can see that they have similar values for the loss in most of the scenarios. Still, for both of the scenarios where a green barrier is present, controller \emph b has a worst result. Controller \emph c has a lower performance in most on the scenarios, where only the first scenario without obstacles is comparable to the other two controllers.

\section{Conclusion}
In this work, we present an approach to characterising relative robustness in image-based robotic control. We present an architecture to train a counterfactual generative model that seeks to produce small, yet close-to-the-original distribution, modifications to a known input during the process of solving classification and regression problems. We evaluate this approach in robot control tasks, after first demonstrating the principle in object classification. Our proposed robustness measure takes into consideration both the controller and the environment, relating it to embodiment theory~\cite{clark1997being}.

\bibliographystyle{IEEEtran}
\bibliography{cgen}

\begin{thebibliography}{10}
\providecommand{\url}[1]{#1}
\csname url@samestyle\endcsname
\providecommand{\newblock}{\relax}
\providecommand{\bibinfo}[2]{#2}
\providecommand{\BIBentrySTDinterwordspacing}{\spaceskip=0pt\relax}
\providecommand{\BIBentryALTinterwordstretchfactor}{4}
\providecommand{\BIBentryALTinterwordspacing}{\spaceskip=\fontdimen2\font plus
\BIBentryALTinterwordstretchfactor\fontdimen3\font minus
  \fontdimen4\font\relax}
\providecommand{\BIBforeignlanguage}[2]{{%
\expandafter\ifx\csname l@#1\endcsname\relax
\typeout{** WARNING: IEEEtran.bst: No hyphenation pattern has been}%
\typeout{** loaded for the language `#1'. Using the pattern for}%
\typeout{** the default language instead.}%
\else
\language=\csname l@#1\endcsname
\fi
#2}}
\providecommand{\BIBdecl}{\relax}
\BIBdecl

\bibitem{zhou1998essentials}
K.~Zhou and J.~C. Doyle, \emph{Essentials of robust control}.\hskip 1em plus
  0.5em minus 0.4em\relax Prentice hall Upper Saddle River, NJ, 1998, vol. 104.

\bibitem{zhang2016understanding}
C.~Zhang, S.~Bengio, M.~Hardt, B.~Recht, and O.~Vinyals, ``Understanding deep
  learning requires rethinking generalization,'' \emph{arXiv preprint
  arXiv:1611.03530}, 2016.

\bibitem{szegedy2013intriguing}
C.~Szegedy, W.~Zaremba, I.~Sutskever, J.~Bruna, D.~Erhan, I.~Goodfellow, and
  R.~Fergus, ``Intriguing properties of neural networks,'' \emph{arXiv preprint
  arXiv:1312.6199}, 2013.

\bibitem{salimans2016improved}
T.~Salimans, I.~Goodfellow, W.~Zaremba, V.~Cheung, A.~Radford, and X.~Chen,
  ``Improved techniques for training {GANs},'' in \emph{Advances in neural
  information processing systems}, 2016, pp. 2234--2242.

\bibitem{heusel2017gans}
M.~Heusel, H.~Ramsauer, T.~Unterthiner, B.~Nessler, and S.~Hochreiter, ``{GAN}s
  trained by a two time-scale update rule converge to a local nash
  equilibrium,'' in \emph{Advances in neural information processing systems},
  2017, pp. 6626--6637.

\bibitem{goodfellow2014generative}
I.~Goodfellow, J.~Pouget-Abadie, M.~Mirza, B.~Xu, D.~Warde-Farley, S.~Ozair,
  A.~Courville, and Y.~Bengio, ``Generative adversarial nets,'' in
  \emph{Advances in neural information processing systems}, 2014, pp.
  2672--2680.

\bibitem{wachter2017counterfactual}
S.~Wachter, B.~Mittelstadt, and C.~Russell, ``Counterfactual explanations
  without opening the black box: Automated decisions and the {GPDR},''
  \emph{Harv. JL \& Tech.}, vol.~31, p. 841, 2017.

\bibitem{glover1992loop}
K.~Glover and D.~McFarlane, ``A loop shaping design procedure using {H}∞-
  synthesis,'' \emph{IEEE Transactions on Automatic Control}, vol.~37, no.~6,
  pp. 759--769, 1992.

\bibitem{la2002design}
M.~La~Civita, G.~Papageorgiou, W.~Messner, and T.~Kanade, ``Design and flight
  testing of a high-bandwidth h-infinity loop shaping controller for a robotic
  helicopter,'' in \emph{AIAA Guidance, Navigation, and Control Conference and
  Exhibit}, 2002, p. 4836.

\bibitem{yeon2008practical}
J.~S. Yeon and J.~H. Park, ``Practical robust control for flexible joint robot
  manipulators,'' in \emph{2008 IEEE international conference on robotics and
  automation}.\hskip 1em plus 0.5em minus 0.4em\relax IEEE, 2008, pp.
  3377--3382.

\bibitem{rigatos2015new}
G.~Rigatos and P.~Siano, ``A new nonlinear h-infinity feedback control approach
  to the problem of autonomous robot navigation,'' \emph{Intelligent Industrial
  Systems}, vol.~1, no.~3, pp. 179--186, 2015.

\bibitem{zinober1989deterministic}
A.~S. Zinober, ``Deterministic control of uncertain systems,'' in
  \emph{Proceedings. ICCON IEEE International Conference on Control and
  Applications}.\hskip 1em plus 0.5em minus 0.4em\relax IEEE, 1989, pp.
  645--650.

\bibitem{yu2005continuous}
S.~Yu, X.~Yu, B.~Shirinzadeh, and Z.~Man, ``Continuous finite-time control for
  robotic manipulators with terminal sliding mode,'' \emph{Automatica},
  vol.~41, no.~11, pp. 1957--1964, 2005.

\bibitem{wang2009neural}
L.~Wang, T.~Chai, and L.~Zhai, ``Neural-network-based terminal sliding-mode
  control of robotic manipulators including actuator dynamics,'' \emph{IEEE
  Transactions on Industrial Electronics}, vol.~56, no.~9, pp. 3296--3304,
  2009.

\bibitem{guldner1995sliding}
J.~Guldner and V.~I. Utkin, ``Sliding mode control for gradient tracking and
  robot navigation using artificial potential fields,'' \emph{IEEE Transactions
  on Robotics and Automation}, vol.~11, no.~2, pp. 247--254, 1995.

\bibitem{zinober1994variable}
A.~S. Zinober, \emph{Variable structure and {L}yapunov control}.\hskip 1em plus
  0.5em minus 0.4em\relax Springer, 1994, vol. 193.

\bibitem{ogren2001control}
P.~Ogren, M.~Egerstedt, and X.~Hu, ``A control {L}yapunov function approach to
  multi-agent coordination,'' in \emph{Proceedings of the 40th IEEE Conference
  on Decision and Control (Cat. No. 01CH37228)}, vol.~2.\hskip 1em plus 0.5em
  minus 0.4em\relax IEEE, 2001, pp. 1150--1155.

\bibitem{park2007performance}
S.~Park, J.~Deyst, and J.~P. How, ``Performance and {L}yapunov stability of a
  nonlinear path following guidance method,'' \emph{Journal of guidance,
  control, and dynamics}, vol.~30, no.~6, pp. 1718--1728, 2007.

\bibitem{ames2014rapidly}
A.~D. Ames, K.~Galloway, K.~Sreenath, and J.~W. Grizzle, ``Rapidly
  exponentially stabilizing control {L}yapunov functions and hybrid zero
  dynamics,'' \emph{IEEE Transactions on Automatic Control}, vol.~59, no.~4,
  pp. 876--891, 2014.

\bibitem{ramamoorthy2006qualitative}
S.~Ramamoorthy and B.~Kuipers, ``Qualitative hybrid control of dynamic bipedal
  walking.'' in \emph{Robotics: Science and Systems.}, 2006.

\bibitem{blight1994practical}
J.~D. Blight, R.~Lane~Dailey, and D.~Gangsaas, ``Practical control law design
  for aircraft using multivariable techniques,'' \emph{International Journal of
  Control}, vol.~59, no.~1, pp. 93--137, 1994.

\bibitem{chu1999tuning}
S.-Y. Chu and C.-C. Teng, ``Tuning of pid controllers based on gain and phase
  margin specifications using fuzzy neural network,'' \emph{Fuzzy sets and
  systems}, vol. 101, no.~1, pp. 21--30, 1999.

\bibitem{sricharan2017semi}
K.~Sricharan, R.~Bala, M.~Shreve, H.~Ding, K.~Saketh, and J.~Sun,
  ``Semi-supervised conditional {GAN}s,'' \emph{arXiv preprint
  arXiv:1708.05789}, 2017.

\bibitem{gatys2016image}
L.~A. Gatys, A.~S. Ecker, and M.~Bethge, ``Image style transfer using
  convolutional neural networks,'' in \emph{Proceedings of the IEEE conference
  on computer vision and pattern recognition}, 2016, pp. 2414--2423.

\bibitem{kingma2013auto}
D.~P. Kingma and M.~Welling, ``Auto-encoding variational {B}ayes,'' \emph{arXiv
  preprint arXiv:1312.6114}, 2013.

\bibitem{mathieu2016disentangling}
M.~F. Mathieu, J.~J. Zhao, J.~Zhao, A.~Ramesh, P.~Sprechmann, and Y.~LeCun,
  ``Disentangling factors of variation in deep representation using adversarial
  training,'' in \emph{Advances in neural information processing systems},
  2016, pp. 5040--5048.

\bibitem{berthelot2018understanding}
D.~Berthelot, C.~Raffel, A.~Roy, and I.~Goodfellow, ``Understanding and
  improving interpolation in autoencoders via an adversarial regularizer,''
  \emph{arXiv preprint arXiv:1807.07543}, 2018.

\bibitem{wu2019imagine}
Y.~Wu, S.~Kasewa, O.~Groth, S.~Salter, L.~Sun, O.~P. Jones, and I.~Posner,
  ``Imagine that! leveraging emergent affordances for tool synthesis in
  reaching tasks,'' \emph{arXiv preprint arXiv:1909.13561}, 2019.

\bibitem{NIPS2017_7152}
\BIBentryALTinterwordspacing
S.~Racani\`{e}re, T.~Weber, D.~Reichert, L.~Buesing, A.~Guez,
  D.~Jimenez~Rezende, A.~Puigdom\`{e}nech~Badia, O.~Vinyals, N.~Heess, Y.~Li,
  R.~Pascanu, P.~Battaglia, D.~Hassabis, D.~Silver, and D.~Wierstra,
  ``Imagination-augmented agents for deep reinforcement learning,'' in
  \emph{Advances in Neural Information Processing Systems 30}, I.~Guyon, U.~V.
  Luxburg, S.~Bengio, H.~Wallach, R.~Fergus, S.~Vishwanathan, and R.~Garnett,
  Eds.\hskip 1em plus 0.5em minus 0.4em\relax Curran Associates, Inc., 2017,
  pp. 5690--5701. [Online]. Available:
  \url{http://papers.nips.cc/paper/7152-imagination-augmented-agents-for-deep-reinforcement-learning.pdf}
\BIBentrySTDinterwordspacing

\bibitem{kalweit2017uncertainty}
G.~Kalweit and J.~Boedecker, ``Uncertainty-driven imagination for continuous
  deep reinforcement learning,'' in \emph{Conference on Robot Learning}, 2017,
  pp. 195--206.

\bibitem{10.1007/978-3-642-28258-4_9}
S.~C. Smith and J.~M. Herrmann, ``Homeokinetic reinforcement learning,'' in
  \emph{Partially Supervised Learning}, F.~Schwenker and E.~Trentin, Eds.\hskip
  1em plus 0.5em minus 0.4em\relax Berlin, Heidelberg: Springer Berlin
  Heidelberg, 2012, pp. 82--91.

\bibitem{lee2018dynamic}
J.~N. Lee, M.~Laskey, A.~K. Tanwani, A.~Aswani, and K.~Goldberg, ``A dynamic
  regret analysis and adaptive regularization algorithm for on-policy robot
  imitation learning,'' in \emph{International Workshop on the Algorithmic
  Foundations of Robotics}.\hskip 1em plus 0.5em minus 0.4em\relax Springer,
  2018, pp. 212--227.

\bibitem{pearl2009causal}
J.~Pearl \emph{et~al.}, ``Causal inference in statistics: An overview,''
  \emph{Statistics surveys}, vol.~3, pp. 96--146, 2009.

\bibitem{pearl2009causality}
J.~Pearl, \emph{Causality}.\hskip 1em plus 0.5em minus 0.4em\relax Cambridge
  university press, 2009.

\bibitem{nair2019causal}
S.~Nair, Y.~Zhu, S.~Savarese, and L.~Fei-Fei, ``Causal induction from visual
  observations for goal directed tasks,'' \emph{arXiv preprint
  arXiv:1910.01751}, 2019.

\bibitem{NIPS2008_3548}
\BIBentryALTinterwordspacing
P.~O. Hoyer, D.~Janzing, J.~M. Mooij, J.~Peters, and B.~Sch\"{o}lkopf,
  ``Nonlinear causal discovery with additive noise models,'' in \emph{Advances
  in Neural Information Processing Systems 21}, D.~Koller, D.~Schuurmans,
  Y.~Bengio, and L.~Bottou, Eds.\hskip 1em plus 0.5em minus 0.4em\relax Curran
  Associates, Inc., 2009, pp. 689--696. [Online]. Available:
  \url{http://papers.nips.cc/paper/3548-nonlinear-causal-discovery-with-additive-noise-models.pdf}
\BIBentrySTDinterwordspacing

\bibitem{shajarisales2015telling}
N.~Shajarisales, D.~Janzing, B.~Sch{\"o}lkopf, and M.~Besserve, ``Telling cause
  from effect in deterministic linear dynamical systems,'' in
  \emph{International Conference on Machine Learning}, 2015, pp. 285--294.

\bibitem{peters2017elements}
J.~Peters, D.~Janzing, and B.~Sch{\"o}lkopf, \emph{Elements of causal
  inference: foundations and learning algorithms}.\hskip 1em plus 0.5em minus
  0.4em\relax MIT press, 2017.

\bibitem{alvarez2017causal}
D.~Alvarez-Melis and T.~S. Jaakkola, ``A causal framework for explaining the
  predictions of black-box sequence-to-sequence models,'' \emph{arXiv preprint
  arXiv:1707.01943}, 2017.

\bibitem{singla2019explanation}
S.~Singla, B.~Pollack, J.~Chen, and K.~Batmanghelich, ``Explanation by
  progressive exaggeration,'' 2019.

\bibitem{kim2017interpretable}
J.~Kim and J.~Canny, ``Interpretable learning for self-driving cars by
  visualizing causal attention,'' in \emph{Proceedings of the IEEE
  international conference on computer vision}, 2017, pp. 2942--2950.

\bibitem{van2019interpretable}
A.~Van~Looveren and J.~Klaise, ``Interpretable counterfactual explanations
  guided by prototypes,'' \emph{arXiv preprint arXiv:1907.02584}, 2019.

\bibitem{goyal2019counterfactual}
Y.~Goyal, Z.~Wu, J.~Ernst, D.~Batra, D.~Parikh, and S.~Lee, ``Counterfactual
  visual explanations,'' \emph{arXiv preprint arXiv:1904.07451}, 2019.

\bibitem{chang2018explaining}
C.-H. Chang, E.~Creager, A.~Goldenberg, and D.~Duvenaud, ``Explaining image
  classifiers by counterfactual generation,'' \emph{arXiv preprint
  arXiv:1807.08024}, 2018.

\bibitem{bongard2006resilient}
J.~Bongard, V.~Zykov, and H.~Lipson, ``Resilient machines through continuous
  self-modeling,'' \emph{Science}, vol. 314, no. 5802, pp. 1118--1121, 2006.

\bibitem{liu2018large}
Z.~Liu, P.~Luo, X.~Wang, and X.~Tang, ``Large-scale celebfaces attributes
  (celeba) dataset,'' \emph{Retrieved August}, vol.~15, p. 2018, 2018.

\bibitem{clark1997being}
A.~Clark and M.~A. Boden, ``Being there: Putting brain, body, and world
  together again,'' \emph{Nature}, vol. 386, no. 6622, pp. 237--237, 1997.

\end{thebibliography}

\end{document}